\pdfoutput=1

\documentclass[11pt]{article}

\usepackage[]{EMNLP2022}

\usepackage{times}
\usepackage{latexsym}

\usepackage{supertabular}
\usepackage[T1]{fontenc}

\usepackage[utf8]{inputenc}

\usepackage{microtype}

\usepackage{inconsolata}

\usepackage{graphicx}

\usepackage{xspace}

\newcommand{\logsdataset}{VIRADialogs}
\newcommand{\systemname}{\textsc{VIRA}\xspace}
\newcommand{\silverintents}{\textsc{Silver labels}\xspace}
\newcommand{\oracleintents}{\textsc{Oracle intents}\xspace}
\newcommand{\predictedintents}{\textsc{Predicted Intents}\xspace}
\newcommand{\predictedoracleintents}{\textsc{Predicted Oracle Intents}\xspace}

\title{Benchmark Data and Evaluation Framework for Intent Discovery Around COVID-19 Vaccine Hesitancy}

 \author {
 Shai Gretz$^1$\thanks{\ \ These authors equally contributed to this work.}, Assaf Toledo$^1$\footnotemark[1],
 Roni Friedman$^1$ \\
 {\bf Dan Lahav$^1$}, {\bf Rose Weeks$^2$}, {\bf Naor Bar-Zeev$^2$} \\
  {\bf João Sedoc$^3$}, {\bf Pooja Sangha$^2$,} {\bf Yoav Katz$^1$}, {\bf Noam Slonim$^1$}\\
 $^1$IBM Research; $^2$Johns Hopkins Bloomberg School of Public Health; $^3$New York University \\

 \
 \{avishaig,roni.friedman-melamed,katz,noams\}@il.ibm.com\\
 \{assaf.toledo,dan.lahav\}@ibm.com, \{rweeks,nbarzee1,psangha1\}@jhu.edu\\
 \{jsedoc\}@stern.nyu.edu \\
}

\begin{document}
\maketitle

\begin{abstract}

The COVID-19 pandemic has made a huge global impact and cost millions of lives. As COVID-19 vaccines were rolled out, they were quickly met with widespread hesitancy. To address the concerns of hesitant people, we launched \systemname, a public dialogue system aimed at addressing questions and concerns surrounding the COVID-19 vaccines.
Here, we release \logsdataset{}, a dataset of over $8k$ dialogues conducted by actual users with \systemname, providing a unique real-world conversational dataset. In light of rapid changes in users' intents, due to updates in guidelines or in response to new information, we highlight the important task of intent discovery in this use-case. We introduce a novel 
automatic evaluation framework for intent discovery, leveraging the existing intent classifier of \systemname.
We use this framework to report baseline intent-discovery results over \logsdataset{}, that highlight the difficulty of this task. 

\end{abstract}

\section{Introduction}

As COVID-19 vaccines became available in late $2020$, they were met with widespread vaccine hesitancy \cite{vaccine_hesitancy_2, vaccine_hesitancy}, a phenomena recognized as a top global concern by the World Health Organization (WHO) in 2019. To address such hesitancy, one needs accurate, reliable, and up to date information, constantly available to the general public.

In recent years, task-oriented Dialogue Systems (DSs) have become an integral part of our daily lives, covering domains such as banking, tourism, and government agencies \cite{chatbots_in_government}.

Correspondingly, we introduced \systemname – the Vaccine Information Resource Assistant 
– an informational DS that aims to engage with COVID-19 vaccination intents.\footnote{Throughout this paper, we use the term \emph{intent} to refer to concise statements or queries, representing a coherent set of utterances; e.g., ``Is the vaccine safe?'', as opposed to a synthetic class name like ``vaccine\_safe''.} \systemname is able to respond to $181$ different intents, accumulated over the course of the pandemic. \systemname responses were written and vetted by leading medical and public health experts, relying on up-to-date facts and guidelines. An example of a dialog conducted with \systemname is presented in Table~\ref{table:dialog_example}.

\begin{table*}
\small
\begin{center}
\begin{tabular}{ |l|p{7cm}|p{5cm}| }
 \hline
 Side & Text & Intent \\
 \hline
 System & \emph{Hey! Ask me your vaccine questions.} & \\
 \hline
  User & \emph{Hi. Should I be concerned about side effects of the vaccine if I'm breastfeeding?} & Is breastfeeding safe with the vaccine \\
 \hline
  System & \emph{Perhaps check this out: while trials did not include breastfeeding moms, COVID-19 vaccines are "non-replicating" vaccines. This type of vaccine poses no risk for breastfed infants, so COVID-19 vaccines are also safe when you're breastfeeding!} & \\
 \hline
  User & \emph{what about if I'm pregnant?} & Does the vaccine impact pregnancy? \\
 \hline
  System & \emph{I understand. That's a really important point! There is evidence that these vaccines are safe in pregnancy. COVID-19 vaccines also do not impair fertility. Because COVID-19 disease is more severe in pregnancy, pregnant women should go get their vaccine!} & \\
   \hline
 \end{tabular}
 \end{center}
 \caption{An example of a dialog in \logsdataset{}.}
\label{table:dialog_example}
\end{table*}

Advancing DSs depends on the availability of conversational datasets on which models can be trained. In domains where fact-based information is a necessity, e.g., healthcare, curating such datasets is often challenging: users in a crowd-sourcing setting may not share their authentic concerns, may not represent public opinion, or may even attempt trolling. In addition, creating responses by highly-trained individuals is a very demanding process \cite{emotional_support}. Furthermore, even if one has collected data from a real-world DS, there could be limitations for making such data public.

The availability of \systemname enabled us to collect dialogs with real-world users, following word-of-mouth or social media advertising, presumably conveying genuine interest or concerns related to the vaccines. \systemname was launched in July $2021$ and over the course of $10$ months it accumulated over $8k$ conversations. We refer to this collection of conversations as \mbox{\logsdataset{}} and release it as part of this work.\footnote{\url{vaxchat.org/research}}

After deploying a DS in a real-world setting, users may introduce new intents, which are not part of the system's predefined intents \cite{chatbots_survey}. \systemname's use case represents such an extreme example where users' intents change rapidly due to updates in guidelines and protocols, or as a response to new information (e.g., the outbreak of novel variants). Hence, we needed to frequently update and expand the set of user intents. This makes \logsdataset{} a unique resource for \emph{Intent discovery} methods. These methods aim to reveal such new intents from conversational logs, trying to identify the most salient new intents, which can then be reviewed and added to the DS using a human-in-a-loop process.

To directly evaluate such methods, one would need to annotate each user utterance with its gold intent, and compare this intent with the prediction of each method, which is typically not feasible in large datasets. As a practical alternative, we propose a novel retrospective evaluation paradigm which leverages the existing intent classifier of \systemname. We assume that this classifier, carefully developed over the entire relevant  
time period, covers most intents present in the data. Thus, we treat it as an Oracle to evaluate various intent discovery methods, independently in each month.

First, the Oracle is used to induce silver labels over the unlabeled user utterances. Next, to evaluate an intent discovery method, the same Oracle is used to classify intents predicted by this method to silver labels, enabling a fully automatic quantitative evaluation. We use this approach to evaluate various intent discovery methods on top of \logsdataset{} and further share the code base to reproduce our experiments.\footnote{\url{https://github.com/IBM/vira-intent-discovery}}

To summarize, the contribution of this paper is three fold: i) We release \logsdataset{}, a unique dataset of real-world human-machine conversations, reflecting COVID-19 vaccine hesitancy; ii) We propose and implement an automatic retrospective evaluation paradigm for intent discovery, relying on the availability of a high quality intent classifier; iii) We use our evaluation approach to report baseline performance of various intent discovery methods on top of \logsdataset{}.

\section{Related Work}

\textbf{Benchmark Datasets and COVID-19 DSs.}
Popular benchmark datasets for intent classification are also used to benchmark the task of intent discovery and were curated by asking crowd-annotators to phrase intents 
suitable to a DS setting (e.g., \citet{XLiu_intent_benchmark, larson-intent_benchmark}). \citet{arora-etal-2020-hint3} introduce HINT3, a challenging benchmark whose test set comes from real chats in $3$ domains. 
However,
the test set contains less than $1{,}000$ queries for each domain collected in a $15$-day period, a 
relatively limited 
scope for intent discovery.

The pandemic outbreak led to the development of a few other DSs
in this domain. \citet{welch-etal-2020-expressive} introduce expressive interviewing -- an interview style aiming 
to encourage users to express their thoughts and feelings  
by asking them questions about how COVID-19 has impacted their lives. \citet{toni_hunter_lisa_covid} built and studied a DS 
specifically addressing COVID-19 vaccine hesitancy and showed that $20\%$ of study participants changed their stance in favor of the vaccine after conversing with the system. While their motivation is similar to ours, the analyzed data
is smaller and coming entirely from crowd annotators.

\textbf{Intent Discovery Methods.} Recent work by \citet{ella-insights} introduced a fully unsupervised pipeline for detecting intents in unhandled DS 
logs. 
Utterances are encoded into vector representations, and a Radius-based Clustering (RBC) algorithm assigns each to an existing cluster, in case it surpasses a predefined similarity threshold; or use it to initiate a new cluster. The algorithm automatically selects the number of clusters, and does not enforce full partitioning of the underlying data, but rather enables outliers — instances that lay in isolation of discovered clusters. The paper also suggests a method for selecting cluster representatives aimed at maintaining centrality and diversity.

Key Point Analysis (KPA) \cite{bar-haim-etal-2020-arguments, bar-haim-etal-2020-quantitative, bar-haim-etal-2021-every} proposes a framework that provides both textual and quantitative summary of the main points in a given data. KPA extracts the main points discussed in a collection of texts, and matches the input sentences to these key points. It has been shown to perform well on argumentative data, as well as in online surveys and on user reviews. To our knowledge, our work is the first to utilize KPA in the context of DSs.

\section{The \systemname System}
\label{sec:virasystem}

Users communicate with \systemname using either 
a WEB-based User Interface (UI)\footnote{\url{vaxchat.org}. 
The UI is also embedded on the web pages of health departments, vaccine advocacy organizations, and health care facilities.} or a WhatsApp application. 
The general flow is that users enter free text expressing their questions and concerns about the vaccine, \systemname detects the intent within  
a pre-defined intent list,
and in turn provides a suitable response,
reviewed by medical experts. \systemname supports conversations in English. Below we describe
\systemname's main components.\\

\textbf{Profanity Classifier}. We 
use a dictionary\footnote{\url{https://github.com/LDNOOBW/}} to 
identify utterances with suspected toxic language,
to which \systemname presents a generic response.

\textbf{Dialog-Act Classifier}.
We classify each user
input to one of the supported dialog acts. For certain dialog acts, 
e.g., `Hi',
\systemname presents a generic response. 
Full details can be found in Appendix~\ref{sec:dialog_act_details}.

\textbf{Intent Classifier}.
Intents representing distinct vaccine concerns were initially identified through various means: using a Twitter analysis, reviewing audience questions in Zoom-based public forums hosted by authors’ affiliated academic centers, and synthesizing web pages with FAQs. Over time, new concerns were identified by monitoring incoming queries to \systemname system and eventually the list comprised of $181$ intents (Appendix~\ref{sec:intentsCovered}).

The intent classifier was trained on data collected from crowd annotators using the Appen platform.\footnote{\url{appen.com}} Annotators were presented with an intent and asked to express it in three different ways, as if conversing with a knowledgeable friend (see Section~\ref{sec:oracle_description} for more details). The classifier's top-ranked intent is selected for providing a response from the Response Database. If no intent passed a pre-defined threshold, a corresponding response is given.

\textbf{Response Database}.
This database contains \systemname's responses to intents. 
Each entry specifies multiple responses to a specific 
intent, to 
increase output diversity.
The responses contain varying information and tone from which \systemname selects one randomly. The database was created and is maintained by experts in the field based on up-to-date facts and guidelines. All responses sought to minimize technical language and maintain brevity through a $280$-character limit.

\textbf{Feedback Mechanism}.
\systemname incorporates a feedback mechanism that enables users
to correct the course of conversation. This feedback allows \systemname's personnel to improve the system over time (see more details in Appendix~\ref{sec:feedback_mechanism_appx}).

All \systemname's chats, including feedback selections and classifiers output, are recorded for off-line analysis, without storing identifiable information.

\section{The \logsdataset{} Dataset}

\logsdataset{} contains the interactions conducted with \systemname by actual users 
from July $2021$ to May $2022$. The full dialogues, as well as user feedback, predicted intents, dialog acts, and toxicity predictions 
are released to the research community. The data has been anonymized by masking 
locations, names, e-mails, phone numbers, and birth-dates, along with suspected toxic terms,
using a range of regular expressions, the Profanity Classifier, and the 
spaCy Named Entity recognizer.\footnote{\url{https://spacy.io/}} In addition, we have excluded dialogues between $29$-$30$, July $2021$,
in which \systemname was confronted with multiple toxic inputs, presumably from individuals who attempted to break the system. Stats of \logsdataset{} are presented in Table~\ref{table:logsStats}.

\begin{table}
\small
\begin{center}
\begin{tabular}{ |l|l|  }
\hline
 \# Dialogs & $8{,}088$\\
 \hline
 Total \# Turns & $28{,}202$\\
 \hline
 Avg. turns per dialog & $3.5$\\
 \hline
 Total \# Turns w/o feedback turns & $20{,}304$ \\
 \hline
 \end{tabular}
 \end{center}
 \caption{Stats of \logsdataset{}. Row 2 includes turns that are both free text and a feedback selection (see Appendix~\ref{sec:feedback_mechanism_appx}), whereas row 4 indicates free text turns only.}
\label{table:logsStats}
\end{table}
\label{sec:logs}

\section{Retrospective Intent Discovery Evaluation}
\label{sec:intent_discovery_silver}

An important contribution of this work is to show how to leverage an existing DS intent {\it classifier\/} -- like the one described in Section~\ref{sec:virasystem}, referred to as an \emph{Oracle} -- to automatically evaluate intent {\it discovery\/} methods over a collection of dialogs. An overview of the proposed approach is depicted in Figure~\ref{fig:oracle_evaluation_paradigm}.
The underlying components are described below, using the following terminology:

\noindent
\begin{figure*}[htb]
\begin{center}
\includegraphics[width=1.7\columnwidth]{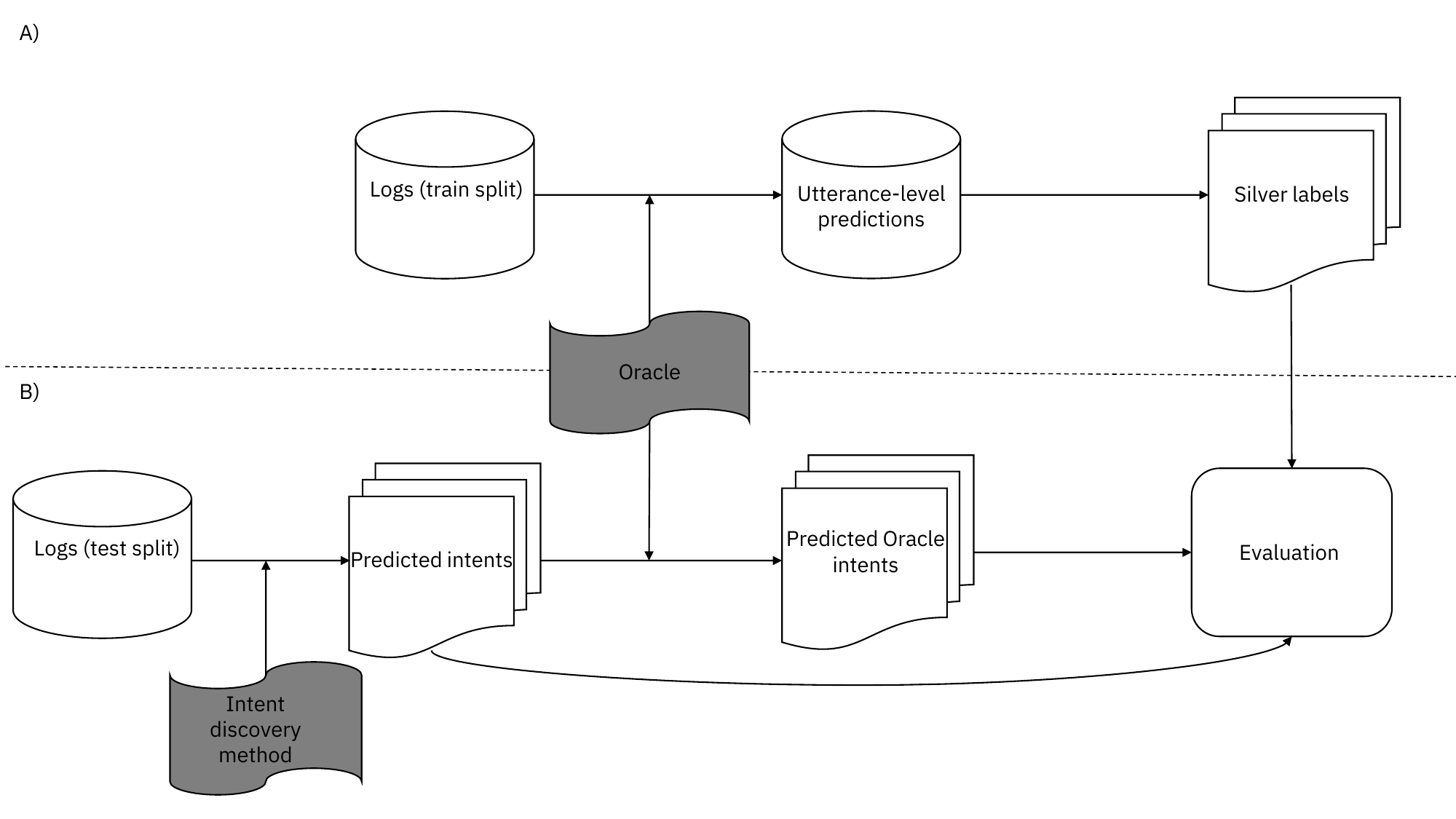}
\caption{Overview of the evaluation pipeline. A) Inducing \silverintents:  (Section~\ref{sec:inducing_silver_labels}): We infer the \emph{Oracle}, an intent classifier, over all utterances in a train split of logs. We rank the obtained clusters of intents and define the top $K$ ones as the \silverintents.
B) Evaluation Method (Section~\ref{sec:system_evaluation}): For \predictedintents of a given intent discovery method, we use the Oracle to classify them to at most one of the \oracleintents, forming clusters of \predictedoracleintents, evaluated w.r.t \silverintents.}
\label{fig:oracle_evaluation_paradigm}
\end{center}
\end{figure*}
\textbf{\oracleintents}: The intents supported by the Oracle.
\textbf{\silverintents}: Subset of \oracleintents, induced over a 
given data.
\textbf{\predictedintents}: Intents predicted and phrased by an
intent discovery method.
\textbf{\predictedoracleintents}: Subset of \predictedintents mapped by the Oracle to \oracleintents.

\subsection{Inducing \silverintents}
\label{sec:inducing_silver_labels}

Given a set of unlabeled user utterances from conversational logs we randomly split it to train and test sets. The train set is used to induce \silverintents, while the test set is used for evaluation.
The motivation of the train-test split is three-fold: (i) enabling to evaluate how consistent is the Oracle itself to ensure the emerging \silverintents are representative of the entire data; (ii) preserving an option to evaluate supervised intent discovery methods in future work; (iii) using the Oracle test set results to estimate upper bound test performance. 

We apply the Oracle to predict the intent of each utterance in the train set. Utterances on which the Oracle confidence was below a pre-specified threshold are placed in a \emph{none} cluster. Next, we sort all \oracleintents based on their predicted prevalence, and define the top $K$ ranked ones as the \silverintents, where ranking criteria can vary. Each of the \silverintents corresponds to a cluster of user utterances mapped to it.

\subsection{Evaluation Method}
\label{sec:system_evaluation}

\subsubsection{Matching \predictedintents to \silverintents}
\label{sec:matching_system_to_silver}

\predictedintents often cannot be matched directly to \silverintents. E.g., an intent discovery method might output ``I don't want to get a booster shot'', whereas the corresponding intent in the \silverintents would be ``Will I need a booster shot?''.
Assuming manual mapping is not feasible, we use the Oracle to map each of the \predictedintents to -- at most -- one of the \oracleintents, resulting in a set of \predictedoracleintents. Utterances of \predictedintents which are not mapped due to low confidence of the Oracle
are placed in a \emph{none} cluster. Note, that in principle this set may contain \oracleintents that were not selected as \silverintents.
In a sense, this mapping normalizes the text associated with \predictedintents to conform with the \silverintents, enabling to evaluate them w.r.t one another.

\subsubsection{Evaluation Measures}
\label{sec:eval_measures}

We consider two types of measures to evaluate intent discovery methods:
(a) the similarity of \predictedintents to \silverintents; and (b) the similarity of cluster partitions generated on the test data by the Oracle and the evaluated method.

\subsubsection*{Intent Discovery Measures}
We estimate the quality of \predictedintents (PIs) using the \predictedoracleintents (POIs) and \silverintents (SLs) as follows:

\textbf{Recall}: $\frac{\textrm{|POIs|}\cap\textrm{|SLs|}} {\textrm{|SLs|}}$ (How many \silverintents did the method cover?) 

\textbf{Precision}: $\frac{\textrm{|POIs|}\cap\textrm{|SLs|}} {\textrm{|PIs|}}$ (How many \predictedintents were mapped to \silverintents?)

\textbf{JS-distance}: We place utterances of \predictedoracleintents not in the \silverintents in the \emph{none} cluster. We normalize the sizes of the clusters induced by the \silverintents and the \predictedoracleintents -- including the \emph{none} cluster -- into two probability distributions, and report their Jensen-Shannon divergence.

\subsubsection*{Intent Clusters' Analysis}
We compare the partitioning of the test data induced by the \predictedintents 
and the Oracle 
using the following standard measures:
\textbf{Adjusted Rand-Index (ARI)}: The rand index corrected for chance \cite{ari-ami1}.
\textbf{Adjusted Mutual-Information (AMI)}: 
The mutual information corrected for chance \cite{ami2}.
\textbf{V-Measure}: The harmonic mean between homogeneity and completeness \cite{vmeasure}.

\section{Experimental Setup}

In this section we present a concrete implementation of the framework
described in Section~\ref{sec:intent_discovery_silver} using \systemname and \logsdataset{} to automatically
evaluate various 
unsupervised
intent discovery methods.

\subsection{The Oracle}
\label{sec:oracle_description}
For the Oracle we use \systemname's intent classifier (Section~\ref{sec:virasystem}), 
described below. 

\subsection*{Data}

For each intent amongst the final $181$ intents covered by \systemname, we asked $18$ Appen crowd annotators to contribute three different \emph{intent expressions}, i.e., different phrasings of questions or comments by which they could have expressed the intent while chatting with a knowledgeable friend.
Qualified annotators were paid on average $7.5$-$8$\$ an hour.\footnote{For each annotator, we calculate the BLEU score of its expressions w.r.t the intent. Annotators with score < $0.07$ are determined as qualified, aiming at promoting diversity.} After manual cleaning we ended up with $7{,}990$ expressions, between $20$-$100$ for each intent.\footnote{The data also contains a small set of $324$ intent expressions, extracted manually from \logsdataset{}.} We release this dataset as part of this work, contributing to the task of single-domain intent classification.\footnote{\url{https://research.ibm.com/haifa/dept/vst/debating_data.shtml}}

\subsection*{Model and Training}

We split the intent expressions associated with each intent to train ($65\%$), dev ($8\%$), and test ($27\%$) sets,
with $5{,}169$, $664$ and $2{,}139$ examples, respectively, over which we fine-tuned RoBERTa-large \cite{liu2019roberta}. Full model implementation details and threshold tuning are in Appendix~\ref{sec:key_point_classification_model}. Note, when the confidence score of the top prediction was below a pre-specified threshold, the model does not predict any intent. 

\subsection{Inducing \silverintents}

As a pre-processing step we filter from \logsdataset{} user input that reflect user feedback, using the feedback mechanism in \systemname, keeping only free text utterances for the following analysis. We also remove utterances longer than $250$ characters, contain at most one non-masked word, or less than $75\%$ alpha-numeric characters. We split the remaining utterances into monthly intervals, resulting in $10$ data folds, and subsequently evenly split the utterances in each fold to train and test (indifferent to the dialogue utterances came from).
 
To reduce noise in generating \silverintents, we additionally filter from the train set utterances classified with a dialog act (e.g., `greeting') or as toxic, as the ratio of intents related to COVID-19 vaccines in these utterances is much smaller.  

We then apply the Oracle on each utterance in the train set, resulting in \oracleintents and corresponding clusters. We sort them based on their prevalence and define the \silverintents by accumulating the clusters until we reach a coverage of $80\%$ (out of all texts on which the Oracle had a confident prediction) 
or cluster size is below $3$.
The number of utterances and resulting \silverintents for each fold are reported in Table~\ref{table:split_stats}.
\begin{table}
\small
\begin{center}
\begin{tabular}{ |p{1cm}|p{0.7cm}|p{0.7cm}|p{1.2cm}| }
\hline
 Fold & Train size & Test size & \# \silverintents \\
 \hline
 Jul-$21$ & $3{,}011$ & $3{,}294$ & $45$\\
 \hline
 Aug-$21$ & $1{,}169$ & $1{,}285$ & $43$\\
 \hline
 Sep-$21$ & $868$ & $911$ & $37$\\
 \hline
 Oct-$21$ & $718$ & $747$ & $34$\\
 \hline
 Nov-$21$ & $506$ & $521$ & $30$\\
 \hline
 Dec-$21$ & $730$ & $769$ & $31$\\
 \hline
 Jan-$22$ & $799$ & $905$ & $40$\\
 \hline
 Feb-$22$ & $239$ & $250$ & $23$\\
 \hline
 Mar-$22$ & $212$ & $229$ & $18$\\
 \hline
 Apr-$22$ & $192$ & $206$ & $20$\\
 \hline
 \end{tabular}
 \end{center}
 \caption{\# utterances in \logsdataset{} splits for intent discovery evaluation. Size is uneven due to additional filtering done on train.}
\label{table:split_stats}
\end{table}
\subsection{Intent Discovery Methods}

\subsubsection{Clustering Algorithms}
\label{sec:traditional_clustering}

We evaluate two clustering algorithms. Since 
one can not assume that
the number of \silverintents is
known {\it a priori\/}, we use $sqrt(N)$ as a simple heuristic to determine the number of clusters, including the \emph{none} cluster, where $N$ is the number of utterances being clustered. Short utterances, containing less than $5$ recognized words, were placed in advance in the \emph{none} cluster. For both clustering algorithms the analysis takes a few minutes on a 
single 
CPU.

\textbf{K-Means.}
We use the K-Means algorithm from the SciKit-Learn package \cite{scikit-learn}
with the default settings,
running it with $10$ random centroid initializations obtained by K-Means++, with
up to $300$ iterations in each run. 
Each utterance was represented using its
Sentence-BERT representation \cite{sbert}.

\textbf{sequential Information Bottleneck (sIB).}
As a strong bag-of-words baseline, we use the sIB algorithm of \citet{sib}.\footnote{\url{https://github.com/IBM/sib}} 
The algorithm uses as input the
Term-Frequency vector representations and is executed with the default settings of $10$ internal random
initializations and up to $15$ iterations in each run. We apply a common pre-processing stage in which stop-words are removed and the remaining words are stemmed. 
  
\subsubsection*{Intent Extraction}

We select a single user utterance per cluster to represent an intent, resulting with the list of \predictedintents. The selection is based on a statistical analysis of n-grams in the data. For each cluster, we first find the n-grams that are significantly more common in this cluster compared to other clusters based on hyper-geometric test ($p=0.05$). 
Then we select the user utterance in the cluster that includes the maximal number of significant n-grams found in that cluster.

\subsubsection{End-to-End Methods}

We evaluate two
end-to-end methods. Both methods are highly parameterized, and for fairness we mostly maintain the default settings without using any labeled data for parameter tuning. They differ from methods described in Section~\ref{sec:traditional_clustering} in two ways: i) They determine the number of clusters internally, and ii) They map utterances to a \emph{none} cluster as they see fit. For comparison purposes, we take the top $sqrt(N)-1$ prevalent clusters for evaluation. The rest of the clusters are added to the \emph{none} cluster.

\textbf{Key Point Analysis (KPA).} We use KPA as provided by the IBM Debater Academic Early Access Program \cite{bar-haim-etal-2021-project}. The underlying model of KPA matches utterances with key point candidates, identified automatically. Utterances
for which no match was found above a threshold are placed in a \emph{none} cluster. Preliminary experiments have shown KPA is producing too few intents, so as an adjustment for this task we: (i) set $limit\_n\_cands=false$ to remove the limit on
the number of key point candidates; (ii) set $n\_top\_kps=1000$ to remove the limit on number of clusters in the output, which also implies no minimal cluster size. The hypothesis is that (i)+(ii) will increase the amount and diversity of resulting key points at the expense of run-time. The KPA service took about $3.5$ hours to complete the analysis.

\textbf{Radius-based Clustering (RBC).} We approached the authors of \citet{ella-insights} to produce the results for this evaluation. As an adjustment, utterances which contain chit-chat and are filtered at the first phase of the algorithm are placed in a \emph{none} cluster. The minimal similarity threshold is set to $0.55$. As with KPA we do not set a minimum size for clusters. RBC took a few minutes to run on a single CPU.

\section{Results and Discussion}

\subsection{The Oracle}
\label{sec:oracle_eval}

We first establish the quality of \systemname's intent classifier used as the Oracle in various ways.

\textbf{Inference on Intent expressions test set}. We evaluate the Oracle on the test set of the collected intent expressions, using the threshold tuned on the dev set
(Section~\ref{sec:oracle_description}). The Oracle achieves a micro-averaged
precision / recall / f1 of $0.85$ / $0.74$ / $0.79$ on dev, and $0.88$ / $0.77$ / $0.82$ on test.

\textbf{Inducing \silverintents and matching \predictedintents.} We evaluate the Oracle's accuracy in (i) inducing \silverintents (Section~\ref{sec:inducing_silver_labels}) and (ii) matching \predictedintents to \silverintents (Section~\ref{sec:matching_system_to_silver}). 

For (i), we randomly sample $10$ \silverintents from the train set of each of the $10$ folds. For each silver label we sample $2$ utterances mapped to it ($200$ < utterance, \silverintents> pairs overall). For half of the pairs, we randomly replace the silver label with one of the other \oracleintents (thus, obtaining negative pairs). We asked $3$ annotators to annotate whether a given pair of texts has a similar intent or meaning, and took the majority vote as the ground-truth (see more details in Appendix~\ref{sec:pair_task_description}). 
The accuracy of the Oracle on this data is $0.85$. 

For (ii), we randomly select from each fold 
and for each evaluated method $5$ pairs of < \predictedintents, \predictedoracleintents> 
where \predictedoracleintents are part of the \silverintents 
($200$ pairs overall).
We use the same annotation task as in (i). The accuracy of the Oracle on these
data is $0.86$.\footnote{1. We note that on average for $24\%$ of \predictedintents the Oracle is not confident, and for an additional $18\%$ the \predictedoracleintents are not part of the \silverintents. 2. For one of the methods there were less than $5$ pairs, so the overall number of pairs is $199$.}

\begin{table}
\small
\begin{center}
\begin{tabular}{ |c|c|c|c| }
\hline
 Recall & Precision & f1 & JS-distance \\
 \hline
 $0.794$ & $0.799$ & $0.795$ & $0.164$\\
 \hline
 \end{tabular}
 \end{center}
 \caption{Evaluation of the Oracle on \logsdataset{} test sets. The numbers are a weighted-average over the monthly intervals. \emph{Takeaway:} The Oracle is reasonably consistent between the train and test sets.}
\label{table:oracle_consistency}
\end{table}

\textbf{Consistency over \logsdataset{} test.} To recall, we evaluate methods on the \emph{test} set w.r.t \silverintents induced from the \emph{train} set. Here, we would like to examine the consistency of the Oracle's predictions between the sets which also implies the representativeness of the \silverintents for the entire data. We do that by inferring the Oracle over the test set of each monthly fold 
to produce clusters around \oracleintents. We then rank them by prevalence and accumulate them to define the \predictedintents (which are also trivially \predictedoracleintents), as was done to induce \silverintents on the train set. The results are presented in Table~\ref{table:oracle_consistency}. The Oracle achieves a weighted-f1 of $0.795$, demonstrating reasonable consistency between the 
train and test split in each fold. This also can be considered an upper limit of success for other methods.

Overall, the above evaluation has shown that the Oracle performs well in matching utterances and \predictedintents to intents, and that \silverintents are relatively representative.

\subsection{Intent Discovery Methods}

\begin{table*}
\small
\begin{center}
\begin{tabular}{ |l|c|c|c|c|c|c|c|c| }
\hline
 & Recall & Precision & f1 & JS-distance & ARI & AMI & Clustering-f1 & V-measure \\
 \hline
 sIB & $0.385$ & $0.523$ & $0.442$ & $0.333$ & $0.045$ & $0.237$ & $0.074$ & $0.368$\\
 \hline
 K-Means & $0.424$ & $0.575$ & $0.485$ & $0.338$ & $0.06$ & $0.319$ & $0.098$ & $0.432$\\
\hline
RBC & $\textbf{0.446}$ & $\textbf{0.605}$ & $\textbf{0.512}$ & $\textbf{0.315}$ & $0.151$ & $0.283$ & $0.194$ & $0.394$ \\
 \hline
 KPA & $0.437$ & $0.568$ & $0.49$ & $0.323$ & $\textbf{0.244}$ & $\textbf{0.38}$ & $\textbf{0.295}$ & $\textbf{0.477}$ \\
\hline
 \end{tabular}
 \end{center}
 \caption{Evaluation of intent discovery methods on \logsdataset{}. The numbers are a weighted-average over the monthly intervals. Best method for each metric is highlighted in bold. \emph{Takeaway:} Methods are able to uncover up to $45\%$ of the intents, demonstrating the difficulty of this task. RBC is able to uncover more intents and at better precision. KPA is much better at uncovering correct placements of utterances within clusters.}
\label{table:system_results}
\end{table*}

Results for the $4$ methods we evaluate are presented in Table~\ref{table:system_results}. RBC has the highest coverage uncovering almost $45\%$ of the \silverintents, and reaching an f1 of $0.512$. These results also indicate the difficulty of this task, as the majority of \silverintents remain undetected. Note that better precision with worse recall, such as with K-Means compared to KPA, suggests more redundancy in the \predictedintents of the former.

KPA is much better at the clustering measures, and is thus useful for finding good examples for each intent. This might be due to KPA's matching engine, trained to match sentences with key points (similarly to intents in \systemname, key points are concise representations of main points in the data).

It should be noted that for simplicity we have used ``off-the-shelf'' methods with minor adaptations, to resemble a real-world setting where a user would like to get a fast impression of how well such methods perform for a given use-case with minimal effort. It is 
likely that with proper tuning of parameters, domain adaptation of underlying models etc., the performance would have been higher.

\begin{figure*}[t]
\includegraphics[width=\textwidth]{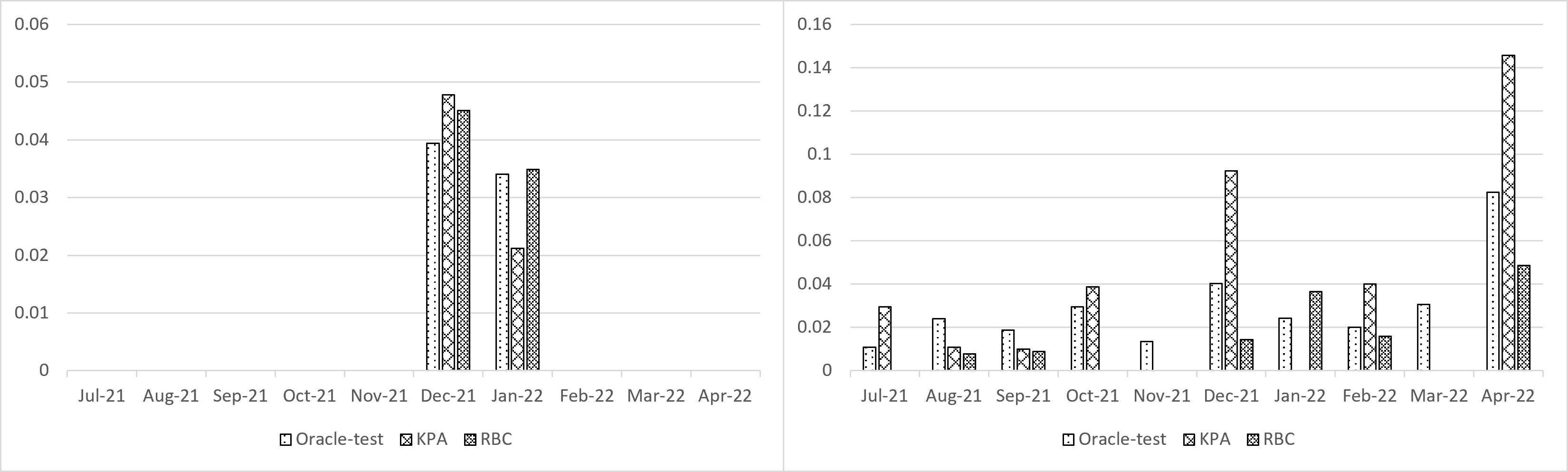}
\centering
\caption{Cluster ratios of \emph{How effective is the vaccine against the Omicron variant} (left); 
\emph{Will I need a booster shot} (right). \emph{Takeaway:} Predictions of methods on \logsdataset{} correlate well with real-world developments.}
\label{fig:main_intents_analysis}
\end{figure*}

\subsection{Qualitative Analysis of Emerging Intents}

The \silverintents and \predictedoracleintents cover varying issues, 
and so we sought to analyze some of the more high-profile ones in light of events that occurred in their context.

We selected two intents: i) \emph{How effective is the vaccine against the Omicron variant}, coupled with the rise in Omicron-related cases in December $2021$;\footnote{\url{https://www.cdc.gov/coronavirus/2019-ncov/science/forecasting/mathematical-modeling-outbreak.html}} and ii) \emph{Will I need a booster shot}, coupled with booster recommendations in late November $2021$\footnote{\url{https://www.cdc.gov/media/releases/2021/s1129-booster-recommendations.html}} and March $2022$.\footnote{{\url{https://www.cdc.gov/media/releases/2022/s0328-covid-19-boosters.html}}} In Figure~\ref{fig:main_intents_analysis}, we plotted the relative cluster size
ratio of each intent among all clusters in a given month, as predicted by the Oracle, KPA, and RBC on the test set. Presumably, high ratio indicates a peak of interest for this intent.

For Omicron, methods highlight emerging interest in December and January, correlated with its real-time occurrence.
To the right, methods predict interest in boosters peaking in December and April.
Overall, this analysis demonstrates how outstanding events in the COVID-19 timeline can be
captured by the evaluated intent discovery methods.

\section{Conclusions}

In this paper we describe \systemname, an informational DS addressing hesitancy towards COVID-19 vaccines. \systemname provides access to accurate, up-to-date information in English, written by experts.
We believe that the associated \logsdataset{} data, containing $8k$ dialogs of \systemname with real-world users, would be a valuable resource to the relevant research community. As an initial example of the potential of this data, we demonstrate how it can be utilized to evaluate intent discovery methods. We propose an automatic evaluation framework that relies on the availability of a corresponding intent classifier, and report the results of $4$ diverse methods, concluding that this benchmark represents a significant challenge.

While automatic evaluation is clearly more practical than manual one, developing the required intent classifier involves a non--trivial effort. Still, we envision two potential outcomes of our work. First, additional intent-discovery methods can be easily evaluated over \logsdataset{} data using our implementation,
and compared to the baseline performance reported here. Second, the same framework can be implemented in other use cases as well for which a reliable intent classifier is available, opening the door for automatic evaluation of intent discovery methods over additional datasets. 

Finally, \systemname is constantly maintained and updated, and is now being expanded to additional languages, to expand its outreach. In future work we intend to report the lessons learned from developing \systemname, and the implications for developing a DS in the public health domain. 

\section*{Acknowledgements}

We thank Ella Rabinovich, Roy Bar-Haim, Yoav Kantor and Lilach Eden for their insightful comments, and Edo Cohen-Karlik, Alex Michel and Elad Venezian for their contribution to \systemname.

\section{Limitations}

There are a few limitations to our approach, which stem from assumptions made to 
establish the evaluation pipeline.

\begin{itemize}
    \item We implement an evaluation pipeline on a single dataset, which we were part of creating, and did not test its compliance with additional datasets.
    \item We assume a relatively accurate intent classifier, referred to as an Oracle, is available. Thus, our evaluation is not suited for cold-start scenarios.
    \item We assume the intents covered by the Oracle indeed cover most intents expressed in the data. It is quite possible that \logsdataset{} included additional intents, beyond the $181$ covered by the Oracle, which probably impacted the accuracy of the evaluation. We note, though, that automatic evaluation, as proposed in this work, is always prone to such issues.
    \item We evaluated only certain unsupervised methods for intent discovery. Other systems may perform better than the reported baselines.
    \item We evaluated only certain unsupervised methods for intent discovery. Other systems, e.g., Watson Assistant Intent Recommendation,\footnote{\url{https://cloud.ibm.com/docs/assistant?topic=assistant-intent-recommendations}} may perform better than the reported baselines.
\end{itemize}

\section{Ethics Statement}

This paper describes work around \systemname, a real-world DS addressing COVID-19 vaccine hesitancy. In an attempt to alleviate concerns that users would take action based on information given to them by \systemname which might harm them, the terms of use of the DS state that ``This information ... is not intended as a substitute for medical advice''. We were guided with the principle of providing accurate information, thus when building \systemname we incorporated a direct mapping between intents and responses. Future endeavours based on this dataset, e.g., for building a generative bot for addressing vaccine hesitancy, should be aware of the ramifications of showing to users such content.

The chats collected might have originally contained offensive language, often as a result of the sensitivity of the domain to some users. We made a dedicated effort to flag these cases and mask problematic terms. However, we did so with automatic measures, so the dataset might still contain such language.

\bibliography{dip_system}
\bibliographystyle{acl_natbib}

\appendix

\section{Dialog-Act Classifier}
\label{sec:dialog_act_details}

This classifier is used for categorizing the user input as one the supported dialog acts: \emph{greeting}, \emph{farewell}, \emph{negative reaction}, \emph{positive reaction}, \emph{concern} and \emph{query}. The classifier was trained on utterances extracted from early chats labeled for their dialog act. \systemname responds to input texts that are classified with one of the first $4$ dialog act types with corresponding generic texts. For example, a response to a greeting (e.g., `Hi') is ``Hello, what are your thoughts about the COVID-19 vaccine?''. Utterances classified as either \emph{concern} or \emph{query} are passed to the Intent Classifier.

\section{Feedback Mechanism}
\label{sec:feedback_mechanism_appx}

\systemname incorporates a feedback mechanism that gives users the option to correct the course of conversation. When users give a thumbs down for a \systemname's response, or when the intent classifier is not confident, \systemname shows to the user the top-$3$ predicted intents in a menu to select from with additional options for indicating that: (a) none of these intents address the concern, or (b) the input does not express a concern at all. This feedback allows \systemname's developers and persons maintaining the Response Database to improve the system over time. For example, when (b) is selected, it indicates a false positive for the Dialog-Act Classifier. 

\section{Intent Classification Model Details}
\label{sec:key_point_classification_model}

As a base model for fine-tuning the intent classifier of \systemname, used as the Oracle, we use RoBERTa-large ($354$M parameters). We use AdamW optimizer with a learning rate of $5$e-$6$ and a batch size of $16$. We fine-tune the model for $15$ epochs and select the best performing checkpoint on the dev set according to overall accuracy. Training took $2.5$ hours on $4$ v$100$ GPUs. The confidence threshold of the model was tuned by taking the minimal threshold such that the precision on the dev set $> 0.85$, resulting in a threshold of $0.296$.



\section{Labeling User Utterances and \predictedintents to \silverintents}
\label{sec:pair_task_description}

We presented annotators with pairs of texts, where one text can be either a user utterance or an intent from the \predictedintents, and the other a silver label. We asked, ``Do the above two texts convey the same meaning or intent?''. The annotators belong to a group with high success on previous tasks of our team, and the task included a few positive and negative examples to illustrate our objective. In addition, we included test questions of text pairs manually selected  from the training data of the Oracle, and annotators with less than $70\%$ accuracy on them were removed from the task.

\section{Intents Supported by \systemname}
\label{sec:intentsCovered}

\tablefirsthead{%
\hline
\multicolumn{1}{|c|}{ Intent}
\\\hline}
\tablehead{%
\hline
\multicolumn{1}{|c|}{ Intent}
\\\hline}
\tabletail{%
\hline}

\small
\begin{supertabular}{|p{7cm}|}

 \hline
 COVID-19 is not as dangerous as they say\\ 
\hline
Do I need to continue safety measures after getting the vaccine?\\ 
\hline
How long until I will be protected after taking the vaccine?\\ 
\hline
How many people already got the vaccine?\\ 
\hline
I am afraid the vaccine will change my DNA\\ 
\hline
I am concerned getting the vaccine because I have a pre-existing condition\\ 
\hline
I am concerned I will be a guinea pig\\ 
\hline
 I'm concerned the vaccine will make me sick.\\ 
\hline
I am not sure if I can trust the government\\ 
\hline
I am young and healthy so I don't think I should vaccinate\\ 
\hline
I distrust this vaccine\\ 
\hline
How much will I have to pay for the vaccine\\ 
\hline
I don't think the vaccine is necessary\\ 
\hline
I don't trust the companies producing the vaccines\\ 
\hline
I don't want my children to get the vaccine\\ 

I think the vaccine was not tested on my community\\ 
\hline
I'm not sure the vaccine is effective enough\\ 
\hline
I'm waiting to see how it affects others\\ 
\hline
COVID vaccines can be worse than the disease itself\\ 
\hline
 Long term side-effects were not researched enough\\ 
\hline
Are regular safety measures enough to stay healthy?\\ 
\hline
Should people that had COVID get the vaccine?\\ 
\hline
Side effects and adverse reactions worry me\\ 
\hline
The COVID vaccine is not safe\\ 
\hline
The vaccine should not be mandatory\\ 
\hline
Do vaccines work against the mutated strains of COVID-19?\\ 
\hline
They will put a chip/microchip to manipulate me\\ 
\hline
What can this chatbot do?\\ 
\hline
What is in the vaccine?\\ 
\hline
Which one of the vaccines should I take?\\ 
\hline
Will I test positive after getting the vaccine?\\ 
\hline
Can other vaccines protect me from COVID-19?\\ 
\hline
Do I qualify for the vaccine?\\ 
\hline
I don't trust vaccines if they're from China or Russia\\ 
\hline
Are the side effects worse for the second shot\\ 
\hline
Can I get a second dose even after a COVID exposure?\\ 
\hline
Can I get other vaccines at the same time?\\ 
\hline
Can I get the vaccine if I have allergies?\\ 
\hline
Can I get the vaccine if I have had allergic reactions to vaccines before?\\ 
\hline
Can I have the vaccine as a Catholic? \\ 
\hline
Can I have the vaccine if I'm allergic to penicillin?\\ 
\hline
Can I still get COVID even after being vaccinated?\\ 
\hline
Can you mix the vaccines?\\ 
\hline
COVID-19 vaccines cause brain inflammation\\ 
\hline
Do the COVID-19 vaccines cause Bell's palsy?\\ 
\hline
"Do the mRNA vaccines contain preservatives, like thimerosal?"\\ 
\hline
Do the vaccines work in obese people?\\ 
\hline
Do you have to be tested for COVID before you vaccinated?\\ 
\hline
Does the vaccine contain animal products?\\ 
\hline
Does the vaccine contain live COVID virus?\\ 
\hline
Does the vaccine impact pregnancy?\\ 
\hline
Does the vaccine work if I do not experience any side effects?\\ 
\hline
How can I stay safe until I'm vaccinated?\\ 
\hline
"How do I know I'm getting a legitimate, authorized vaccine?"\\ 
\hline
How do I report an adverse reaction or side-effect \\ 
\hline
How long do I have to wait between doses?\\ 
\hline
How many doses do I need?\\ 
\hline
How was the vaccine tested?\\ 
\hline
I am concerned about getting the vaccine because of my medications.\\ 
\hline
I don't want the v-safe app monitoring or tracking me\\ 
\hline
I don't want to share my personal information \\ 
\hline
Is breastfeeding safe with the vaccine\\ 
\hline
Is the Johnson \& Johnson vaccine less effective than the others?\\ 

\hline
Is the vaccine halal? \\ 
\hline
Is the vaccine Kosher?\\ 
\hline
Is there vaccine safety monitoring?\\ 
\hline
Other vaccines have caused long-term health problems\\ 
\hline
Should I get the COVID-19 vaccine if I am immunocompromised\\ 
\hline
Should I get the vaccine if I've tested positive for antibodies?\\ 
\hline
The vaccine includes fetal tissue or abortion by-products\\ 
\hline
The vaccine was rushed\\ 
\hline
Vaccine side effects are not getting reported\\ 
\hline
What does vaccine efficacy mean?\\ 
\hline
What if I still get infected even after receiving the vaccine?\\ 
\hline
What if I've been treated with convalescent plasma? \\ 
\hline
What if I've been treated with monoclonal antibodies?\\ 
\hline
What is mRNA?\\ 
\hline
What is the difference between mRNA and viral vector vaccines? \\ 
\hline
When can I go back to normal life? \\ 
\hline
Why are there different vaccines?\\ 
\hline
Why do I need the COVID vaccine if I don't get immunized for flu\\ 
\hline
Why do we need the vaccine if we can wait for herd immunity?\\ 
\hline
Why get vaccinated if I can still transmit the virus?\\ 
\hline
Will 1 dose of vaccine protect me?\\ 
\hline
Can I take a pain reliever when I get vaccinated?\\ 
\hline
Will the vaccine benefit me?\\ 
\hline
Will the vaccine make me sterile or infertile?\\ 
\hline
Can we change the vaccine quickly if the virus mutates?\\ 
\hline
Can I get COVID-19 from the vaccine?\\ 
\hline
I'm still experiencing COVID symptoms even after testing negative - should I still take the vaccine?\\ 
\hline
 Can children get the vaccine?\\ 
\hline
 Can we choose which vaccine we want?\\ 
\hline
 How long does the immunity from the vaccine last?\\ 
\hline
" The mortality rate of COVID-19 is low, why should I get the vaccine?"\\ 
\hline
 There are many reports of severe side effects or deaths from the vaccine\\ 
\hline
How can I get the vaccine?\\ 
\hline
I am worried about blood clots as a result of the vaccine\\ 
\hline
what is covid?\\ 
\hline
Who developed the vaccine?\\ 
\hline
Which vaccines are available?\\ 
\hline
What are the side effect of the vaccine?\\ 
\hline
Can I meet in groups after I'm vaccinated? \\ 
\hline
Is it safe to go to the gym indoors if I'm vaccinated? \\ 
\hline
How do I protect myself indoors? \\ 
\hline
What are the effects of long COVID?\\ 
\hline
Do you need a social security number to get a COVID-19 vaccine? \\ 
\hline
Do you need to be a U.S. citizen to get a COVID-19 vaccine? \\ 
\hline
Is it okay for me to travel internationally if I'm vaccinated? \\ 

\hline
Can my kids go back to school without a vaccine?\\ 
\hline
Will I need a booster shot? \\ 
\hline
"If I live with an immuno-compromised individual, do I still need to wear a mask outdoors if I'm vaccinated? "\\ 
\hline
Does the vaccine prevent transmission? \\ 
\hline
Why is AstraZeneca not approved in the USA?\\ 
\hline
Do I need to change my masking and social distancing practices depending on which COVID-19 vaccine I got? \\ 
\hline
Does the Pfizer vaccine cause myocarditis? \\ 
\hline
Does the Pfizer vaccine cause heart problems? \\ 
\hline
What can you tell me about COVID-19 vaccines?\\ 
\hline
Are there medical contraindications to the vaccines?\\ 
\hline
How many people died from COVID-19?\\ 
\hline
What about reports of abnormal periods due to the vaccine?\\ 
\hline
Do I need the vaccine?\\ 
\hline
Tell me about the vaccine\\ 
\hline
Is the Pfizer vaccine safe for young men?\\ 
\hline
Will vaccination lead to more dangerous variants?\\ 
\hline
Is it safe for my baby to get the vaccine?\\ 
\hline
Did a volunteer in the Oxford trial die?\\ 
\hline
Can I get COVID-19 twice?\\ 
\hline
Are some vaccines safer for younger children than others?\\ 
\hline
How long am I immune from COVID-19 if I had the virus?\\ 
\hline
Are women more likely to get worse side effects than men?\\ 
\hline
How do I convince my family and friends to get the COVID-19 vaccine?\\ 
\hline
Why are COVID-19 vaccination rates slowing in the U.S.?\\ 
\hline
I'm going to get vaccinated\\ 
\hline
Is getting vaccinated painful?\\ 
\hline
What do I do if I lose my COVID-19 vaccination card?\\ 
\hline
Can I get swollen lymph nodes from the vaccine?\\ 
\hline
Can my newborn become immune to COVID-19 if I'm vaccinated?\\ 
\hline
"COVID-19 is over, why should I get the vaccine?"\\ 
\hline
Did one woman die after getting the J\&J vaccine?\\ 
\hline
Do people become magnetic after getting vaccinated?\\ 
\hline
Does the vaccine contain eggs?\\ 
\hline
How is the COVID-19 vaccine different than others?\\ 
\hline
How soon after I've had COVID-19 can I get the vaccination?\\ 
\hline
Is it safe for my teen to get the vaccine?\\ 
\hline
Is this Pfizer vaccine equally effective in kids as it is in adults?\\ 
\hline
Were the COVID-19 vaccines tested on animals?\\ 
\hline
What are the side effects of the vaccine in children?\\ 
\hline
What is the delta variant?\\ 
\hline
What is the J\&J vaccine?\\ 
\hline
What is the Moderna vaccine?\\ 
\hline
What is the Pfizer vaccine?\\ 
\hline
Where are we required to wear masks now?\\ 
\hline
Who can get the Pfizer vaccine?\\ 
\hline
Who can I talk to about COVID-19 in person?\\

\hline
Why should I trust you?\\ 
\hline
Will my child need my permission to get vaccinated?\\ 
\hline
Will the US reach herd immunity?\\ 
\hline
Will my child miss school when they get vaccinated?\\ 
\hline
Is the vaccine FDA approved?\\ 
\hline
Why do vaccinated people need to wear a mask indoors?\\ 
\hline
Do vaccinated people need to quarantine if exposed to COVID-19?\\ 
\hline
What is Ivermectin?\\ 
\hline
Does the Johnson and Johnson vaccine cause Rare Nerve Syndrome?\\ 
\hline
What is the difference between quarantine and isolation?\\ 
\hline
Does the COVID-19 vaccine cause autism?\\ 
\hline
Does the vaccine cause impotence?\\ 
\hline
Who is required to get vaccinated under the federal vaccine mandate?\\ 
\hline
Is the Delta variant more dangerous for kids?\\ 
\hline
Will there be a booster shot for J\&J and Moderna?\\ 
\hline
Is the booster the same as the original vaccine?\\ 
\hline
What are the side effects of booster shots?\\ 
\hline
What is the difference between the third shot and a booster shot?\\ 
\hline
How common are vaccine side effects?\\ 
\hline
Why do my kids need a vaccine if they're unlikely to get sick with COVID-19?\\ 
\hline
What happens if there is a COVID-19 case at my child's school?\\ 
\hline
Are booster shot side effects worse than those from the second shot?\\ 
\hline
Is the booster shot dangerous?\\ 
\hline
Can I get the vaccine if I have Multiple Sclerosis?\\ 
\hline
Do children receive the same dose of Pfizer as adults?\\ 
\hline
What is the Omicron variant?\\ 
\hline
How effective is the vaccine against the Omicron variant?\\
\hline
 \end{supertabular}

\end{document}